OXFORD

# Modeling Polypharmacy Side Effects with Graph Convolutional Networks

**Marinka Zitnik** [1], **Monica Agrawal** [1] and **Jure Leskovec** [1,2,*]

[1]Department of Computer Science, Stanford University, Stanford, CA, USA
[2]Chan Zuckerberg Biohub, San Francisco, CA, USA

*To whom correspondence should be addressed.

## Abstract

**Motivation:** The use of drug combinations, termed polypharmacy, is common to treat patients with complex diseases or co-existing conditions. However, a major consequence of polypharmacy is a much higher risk of adverse side effects for the patient. Polypharmacy side effects emerge because of drug-drug interactions, in which activity of one drug may change, favorably or unfavorably, if taken with another drug. The knowledge of drug interactions is often limited because these complex relationships are rare, and are usually not observed in relatively small clinical testing. Discovering polypharmacy side effects thus remains an important challenge with significant implications for patient mortality and morbidity.

**Results:** Here, we present *Decagon*, an approach for modeling polypharmacy side effects. The approach constructs a multimodal graph of protein-protein interactions, drug-protein target interactions, and the polypharmacy side effects, which are represented as drug-drug interactions, where each side effect is an edge of a different type. *Decagon* is developed specifically to handle such multimodal graphs with a large number of edge types. Our approach develops a new graph convolutional neural network for multirelational link prediction in multimodal networks. Unlike approaches limited to predicting simple drug-drug interaction values, *Decagon* can predict the exact side effect, if any, through which a given drug combination manifests clinically. *Decagon* accurately predicts polypharmacy side effects, outperforming baselines by up to 69%. We find that it automatically learns representations of side effects indicative of co-occurrence of polypharmacy in patients. Furthermore, *Decagon* models particularly well polypharmacy side effects that have a strong molecular basis, while on predominantly non-molecular side effects, it achieves good performance because of effective sharing of model parameters across edge types. *Decagon* opens up opportunities to use large pharmacogenomic and patient population data to flag and prioritize polypharmacy side effects for follow-up analysis via formal pharmacological studies.

**Availability:** Source code and preprocessed datasets are at: http://snap.stanford.edu/decagon.
**Contact:** jure@cs.stanford.edu

## 1 Introduction

Most human diseases are caused by complex biological processes that are resistant to the activity of any single drug (Jia *et al.*, 2009; Han *et al.*, 2017). A promising strategy to combat diseases is polypharmacy, a type of combinatorial therapy that involves a concurrent use of multiple medications, also termed a drug combination (Bansal *et al.*, 2014). A drug combination consists of multiple drugs, each of which has generally been used as a single effective medication in a patient population. Since drugs in a drug combination can modulate the activity of distinct proteins, drug combinations can improve therapeutic efficacy by overcoming the redundancy in underlying biological processes (Sun *et al.*, 2015). For example, a drug combination of Venetoclax and Idasanutlin has recently been shown to lead to superior antileukemic efficacy in the treatment of acute myeloid leukemia (Pan *et al.*, 2017). Here, the two drugs work in reciprocal ways: Venetoclax inhibits antiapoptotic Bcl-2 family proteins while Idasanutlin activates the p53 pathway, and therefore, the combination of these two drugs improves survival by simultaneously targeting complementary mechanisms (Pan *et al.*, 2017).

While the use of multiple drugs may be a good practice for the treatment of many diseases (Liebler and Guengerich, 2005; Tatonetti *et al.*, 2012), a major consequence of polypharmacy to a patient is a much higher risk of side effects which are due to drug-drug interactions. *Polypharmacy side effects* are difficult to identify manually because they are rare, it is practically impossible to test all possible pairs of drugs, and side effects are usually not observed in relatively small clinical testing (Tatonetti *et al.*, 2012; Bansal *et al.*, 2014). Furthermore, polypharmacy is recognized as an increasingly serious problem in the health care system affecting nearly 15% of the U.S. population (Kantor *et al.*, 2015), and costing more than $177 billion a year in the U.S. in treating polypharmacy side effects (Ernst and Grizzle, 2001).

*In vitro* experiments and clinical trials can be performed to identify drug-drug interactions (Li *et al.*, 2015; Ryall and Tan, 2015), but systematic combinatorial screening of drug-drug interaction candidates remains challenging and expensive (Bansal *et al.*, 2014). Researchers have thus attempted to collect drug-drug interactions from scientific literature and electronic medical records (Percha *et al.*, 2012; Vilar *et al.*, 2017), and also discovered them through network modeling, analysis of molecular target signatures (Sun *et al.*, 2015; Huang *et al.*, 2014b; Lewis *et al.*, 2015; Chen *et al.*, 2016a; Takeda *et al.*, 2017), statistical association-based models, and semi-supervised learning (Zhao *et al.*, 2011; Huang *et al.*, 2014a; Chen *et al.*, 2016b; Shi *et al.*, 2017) (see related work in Section 7). While these approaches can be useful to derive broad rules for describing drug interaction at the cellular level, they cannot directly guide strategies for drug combination treatments. In particular, these approaches characterize drug-drug interactions through scores representing the overall probability/strength of an interaction but cannot predict the exact type of the side effect. More precisely, for drugs $i$ and $j$ these methods predict if their combination produces any exaggerated response $S_{ij}$ over and beyond the additive response expected under no interaction, regardless of the exact type or the number of side effects. That is, their goal is to answer a question: $S_{ij} \overset{?}{\neq} \{\}$, where $S_{ij}$ is the set of all polypharmacy side effects attributed specifically to a drug pair $i$, $j$ but not to either drug alone. However, it is much more important and useful to answer whether a pair of drugs $i$, $j$ will interact with a given side effect of type $r$, $r \overset{?}{\in} S_{ij}$. Even though identification of precise polypharmacy side effects is critical for improved patient care, it remains a challenging task that has not yet been studied through predictive modeling.

**Present work.** Here, we develop *Decagon*, a method for predicting side effects of drug pairs. We model the problem by constructing a large two-layer multimodal graph of protein-protein interactions, drug-protein interactions, and drug-drug interactions (*i.e.*, side effects) (Figure 1). Each drug-drug interaction is labeled by a different edge type, which signifies the type of the side effect. We then develop a new multirelational edge prediction model that uses the multimodal graph to predict drug-drug interactions as well as their types. Our model is a convolutional graph neural network that operates in a multirelational setting.

To motivate our model, we first perform exploratory analysis leading to two important observations (Section 3). First, we find that co-prescribed drugs (*i.e.*, drug combinations) tend to have more target proteins in common than random drug pairs, suggesting that drug-target protein information contains valuable information for drug combination modeling. Second, we find that it is important to consider a map of protein-protein interactions in order to be able to model characteristics of drugs with common side effects. These observations motivate the development of *Decagon* to make predictions about which drug pairs will interact and what will the exact type of the interaction/side effect be (Section 4).

*Decagon* develops a new graph auto-encoder approach (Hamilton *et al.*, 2017a), which allows us to develop an end-to-end trainable model for link prediction on a multimodal graph. In contrast, previous graph-based approaches for link prediction tasks in biology (*e.g.*, Huang *et al.* (2014b); Chen *et al.* (2016b); Zong *et al.* (2017)) employ a two-stage pipeline, typically consisting of a graph feature extraction model and a link prediction model, both of which are trained separately. Furthermore, the crucial distinguishing characteristic of *Decagon* is the *multirelational link prediction* ability allowing us to capture the interdependence of different edge (side effect) types, and to identify which out of all possible edge types exist between any two drug nodes in the graph. This is in sharp contrast with approaches for simple link prediction (Trouillon *et al.*, 2016), which predict only existence of edges between node pairs, and is also critical for modeling a large number of different edge/side effect types.

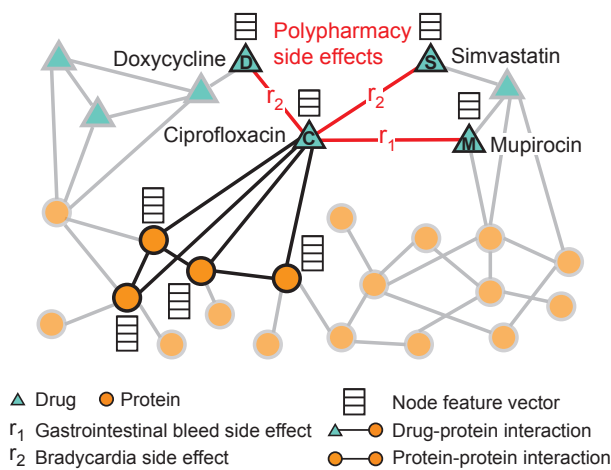

**Fig. 1.** An example graph of polypharmacy side effects derived from genomic and patient population data. A multimodal graph consists of protein-protein interactions, drug-protein targets, and drug-drug interactions encoded by 964 different polypharmacy side effects (*i.e.*, edge types $r_i$, $i = 1, \dots, 964$). Side information is integrated into the model in the form of additional protein and drug feature vectors. Highlighted network neighbors of Ciprofloxacin (node $C$) indicate this drug targets four proteins and interacts with three other drugs. The graph encodes information that Ciprofloxacin (node $C$) taken together with Doxycycline (node $D$) or with Simvastatin (node $S$) increases the risk of bradycardia side effect (side effect type $r_2$), and its combination with Mupirocin ($M$) increases the risk of gastrointestinal bleed side effect $r_1$. We use the graph representation to develop Decagon, a graph convolutional neural model of polypharmacy side effects. Decagon predicts associations between pairs of drugs and side effects (shown in red) with the goal of identifying side effects, which cannot be attributed to either individual drug in the pair.

We contrast *Decagon*'s performance with that of state-of-the-art approaches for multirelational tensor factorization (Nickel *et al.*, 2011; Papalexakis *et al.*, 2017), approaches for representation learning on graphs (Perozzi *et al.*, 2014; Zong *et al.*, 2017), and established machine learning methods for link prediction, which we adapted for the polypharmacy side effect prediction task. *Decagon* outperforms alternative approaches by up to 69% and leads to a 20% average gain in predictive performance, with larger gains achieved on side effect types that have a strong molecular basis (Section 6). For several novel predictions we find supporting evidence in the biomedical literature, suggesting that *Decagon* performs especially well at identifying predictions that are highly likely to be true positive. Taken together, this study shows, for the first time, the ability to model side effects of drug combinations and opens up new opportunities for development of combinatorial drug therapies.

## 2 Datasets

We formulate the polypharmacy side effect identification problem as a multirelational link prediction problem in a two-layer multimodal graph/network of two node types: drugs and proteins. We construct two-layer multimodal network as follows (Figure 1). Protein-protein interaction network describes relationships between proteins. Drug-drug interaction network contains 964 different types of edges (one for each side effect type) and describes which drug pairs lead to which side effects. Lastly, drug-protein links describe the proteins targeted by a given drug.

We continue by describing the datasets used to construct the network. Preprocessed versions of all datasets are available through this study's website: http://snap.stanford.edu/decagon.

## 2.1 Protein-protein and drug-protein interactions

We used the human protein-protein interaction (PPI) network compiled by Menche *et al.* (2015) and Chatr-Aryamontri *et al.* (2015), integrated with additional PPI information from Szklarczyk *et al.* (2017), and Rolland *et al.* (2014). The network contains physical interactions experimentally documented in humans, such as metabolic enzyme-coupled interactions and signaling interactions. The network is unweighted and undirected with 19,085 proteins and 719,402 physical interactions.

We obtained relationships between proteins and drugs from the STITCH (Search Tool for InTeractions of CHemicals) database, which integrates various chemical and protein networks (Szklarczyk *et al.*, 2015). For this work, we considered only the interactions between small chemicals (*i.e.*, drugs) and target proteins that had been experimentally verified. There were over 8,083,600 interactions present between 8,934 proteins and 519,022 chemicals.

## 2.2 Drug-drug interaction and side effect data

We also pulled from databases detailing side effects of both individual drugs and drug combinations. The SIDER (Side Effect Resource) database contains 286,399 drug-side effect associations over 1,556 drugs and 5,868 side effects (Kuhn *et al.*, 2015) obtained by mining adverse events from drug label text. We integrated it with the OFFSIDES database, which details off-label 487,530 associations between 1,332 drugs and 10,097 side effects (Tatonetti *et al.*, 2012). The OFFSIDES database was generated using adverse event reporting systems that collect reports from doctors, patients, and drug companies. We eliminated side effect synonyms and used one side effect vocabulary to construct all datasets. That preprocessing is important as the prediction problem would be much easier if some side effects were perfectly correlated. After combining these datasets, there is a median of 159 side effects per drug, with the most common side effects being nausea, vomiting, headache, diarrhoea, and dermatitis.

We pulled polypharmacy side effect information from TWOSIDES, which details 1,318 side effects types across 63,473 drug combinations, which are greater than expected given the effect of either drug in the combination individually (Tatonetti *et al.*, 2012). Like OFFSIDES, TWOSIDES was generated from adverse event reporting systems. Common side effects, like hypotension and nausea, occur in over a third of drug combinations, while others like amnesia and muscle spasms only occur in a handful of drug combinations. Overall, it contains 4,651,131 drug combination-side effect associations. In this work, we focus on predicting the 964 commonly-occurring types of polypharmacy side effects that each occurred in at least 500 drug combinations.

The final network after linking entity vocabularies used by different databases has 645 drug and 19,085 protein nodes connected by 715,612 protein-protein, 4,651,131 drug-drug, and 18,596 drug-protein edges.

## 3 Data-driven motivation for *Decagon* approach

Here we make three observations about the structure of the two-layer multimodal graph (Figure 1) that have important implications for the design of the *Decagon* model.

First, we observe that there is a wide range in how frequently certain side effects occur in drug combinations. We find that more than 53% of polypharmacy side effects are known to occur in less than 3% of the documented drug combinations (*e.g.*, cerebral artery embolism, lung abscess, sarcoma, collagen disorder). In contrast, the more frequent side effects, (*e.g.*, vomiting, weight gain, nausea, anaemia), occur an order of magnitude more often. Due to the large variation in the number of drug pairs each side effect is associated with, there are only a limited number of drug pairs available for independently training models for prediction of different side effect types. As a result, polypharmacy side-effect prediction becomes a challenging task, especially when predicting rarer side effects, and thus it is important to develop an end-to-end approach such that the model is able to share information and learn from all side effects at once.

Table 1. Percent co-occurrence of hypertension and nausea with the 50 most frequent side effects in drug combinations, annotated with examples. The vast majority of side effects are either significantly overrepresented or underrepresented with respect to how often they appear in drug combinations with nausea/hypertension, at $\alpha = 0.05$, after Bonferroni correction.

| Polypharmacy side effect $S$ | Overrepresented co-occurrence | Underrepresented co-occurrence | Insignificant co-occurrence |
|---|---|---|---|
| Hypertension | 44% (hyperglycemia, anxiety, dizziness) | 48% (fever, sepsis, dermatitis) | 8% (cough, tachycardia) |
| Nausea | 54% (diarrhea, insomnia, asthenia) | 34% (edema, anemia, neutropenia) | 12% (fever, dyspnea) |

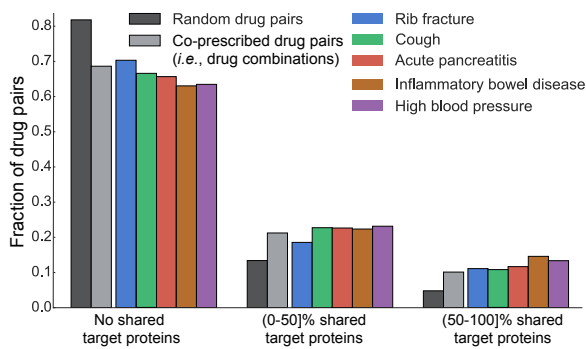

**Fig. 2.** Jaccard similarity between target proteins for random pairs of drugs, all drug pairs, and drug combinations associated with specific side effects. Drug pairs are stratified into three groups depending on whether drug $i$ and $j$ in a given pair $(i, j)$ do not share any target proteins, share fewer than 50% target proteins, or share more than 50% target proteins (i.e., Jaccard($T_i$, $T_j$) = 0, 0 < Jaccard($T_i$, $T_j$) < 0.5, and $0.5 \leq$ Jaccard($T_i$, $T_j$) $\leq 1$, respectively; $T_i$ is a set of $i$'s target proteins). We observe that drugs in most drug pairs, especially in random drug pairs (i.e., drugs not commonly co-prescribed, dark grey) have zero shared target proteins.

Second, we observe that polypharmacy side effects do not appear independently of one another in co-prescribed drug pairs (*i.e.*, drug combinations), suggesting that joint modeling over multiple side effects can aid in the prediction task. To quantify the co-occurrence between side effects, we count the number of drug combinations in which a given side effect co-occurs with other side effects, and then use permutation testing with a null model of random co-occurrence. As exemplified for hypertension and nausea in Table 1, we find that the majority of the most common side effects are either significantly overrepresented or underrepresented with respect to how often they co-occur with nausea/hypertension as side effects in drug combinations, at $\alpha = 0.05$. This observation points to the existence of mechanisms that may contribute to the shared pathophysiology of side effects, similar to what has been observed in disease comorbidity (Lee *et al.*, 2008). For example, we find that hypertension significantly co-occurs with anxiety but co-occurs less often with fever than dictated by random chance (Table 1). These relationships hold across the side effect data set. We conclude that a prediction model should leverage dependence between side effects and be able to re-use the information learned about the molecular basis of one side effect to better understand the molecular basis of another side effect.

Third, we probe the relationship between proteins targeted by a drug pair and occurrence of side effects. Let $T_i$ represent a set of target proteins associated with drug $i$, we then calculate the Jaccard similarity between target proteins of a given drug pair $(i, j)$. We make several observations: (1) More than 68% of drug combinations have zero target proteins in common, suggesting it is important to use protein-protein interaction information to "connect" different proteins targeted by different drugs. (2) Random drug pairs have smaller overlap in targeted proteins than co-prescribed drugs (Figure 2, light grey), p-value = 5e−120, 2-sample Kolmogorov-Smirnov (KS) test. (3) We find that this trend is unequally observed across different side effects. For example, high blood pressure more strongly appears in drug combinations with shared target proteins than, for example, rib fracture (Figure 2, purple). Over 150 side effects appear in combinations that differ significantly (at $\alpha = 0.05$ after Bonferroni correction) from the other true drug combinations, per a 2-sample KS test, suggesting a strong molecular basis of these side effects. Based on this findings, we conclude it is important for a model to consider how proteins interact with each other and to be able to model longer chains of (indirect) interactions.

## 4 Graph convolutional *Decagon* approach

We cast polypharmacy side effect modeling as a *multirelational link prediction problem on a multimodal graph* encoding drug, protein, and side effect relationships (Figure 1). More precisely, these relationships are represented by a graph $G = (\mathcal{V}, \mathcal{R})$ with $N$ nodes (*e.g.*, proteins, drugs) $v_i \in \mathcal{V}$ and labeled edges (relations) $(v_i, r, v_j)$, where $r$ is the edge type (relation type): (1) physical binding between two proteins, (2) a target relationship between a drug and a protein, or (3) a particular type of a side effect between two drugs. As mentioned in Section 2 we consider 964 different relation types between drugs (*i.e.*, side effects).

In addition, we allow for inclusion of side information in the form of additional node features. Different nodes (drugs, proteins) can have different number of node features, given by real-valued feature vectors $\mathbf{x}_1, \mathbf{x}_2, \ldots, \mathbf{x}_N$ assigned to every node in the graph.

**Polypharmacy side effect prediction task.** Polypharmacy side effect prediction task considers the problem of identifying associations between drug pairs and side effects. Importantly, these associations are limited to only those that cannot be attributed to either drug alone. Using the graph $G$, the task is to predict labeled edges between drug nodes. Given a drug pair $(v_i, v_j)$, our aim is to determine how likely an edge $e_{ij} = (v_i, r, v_j)$ of type $r$ belongs to $\mathcal{R}$, meaning that *concurrent* use of drugs $v_i$ and $v_j$ (*i.e.*, the use of a drug combination $(v_i, v_j)$) is associated with a polypharmacy side effect of type $r$ in the human patient population.

To this aim, we develop a non-linear, multi-layer convolutional graph neural network model *Decagon* that operates directly on graph $G$. *Decagon* has two main components:

- *an encoder:* a graph convolutional network operating on $G$ and producing embeddings for nodes in $G$ (Figure 3A) (Section 4.1), and
- *a decoder:* a tensor factorization model using these embeddings to model polypharmacy side effects (Figure 3B) (Section 4.2).

We proceed by describing *Decagon*, our approach for modeling polypharmacy side effects.

### 4.1 Graph convolutional encoder

We first describe the graph encoder model, which takes as input a graph $G$ and additional node feature vectors $\mathbf{x}_i$, and produces a node $d$-dimensional embedding $\mathbf{z}_i \in \mathbb{R}^d$ for every node (drug, protein) in the graph.

We propose an encoder model that makes efficient use of information sharing across regions in the graph and assigns separate processing channels for each relation type. The idea is that *Decagon* learns how to transform and propagate information, captured by node feature vectors, across the graph. Every node's network neighborhood defines a different neural network information propagation architecture but these architectures then share functions/parameters that define how information is shared and propagated. We learn convolutional operators that propagate and transform information across different parts of the graph and across different relation types. The model inspired by a recent class of convolutional neural networks that operate directly on graphs (Defferrard *et al.*, 2016; Kipf and Welling, 2016). For a given node *Decagon* performs transformation/aggregation operations on feature vectors of its neighbors. This way *Decagon* only takes into account the first-order neighborhood of a node and applies the same transformation across all locations in the graph. Successive application of these operations then effectively convolves information across the $K$-th order neighborhood (*i.e.*, embedding of a node depends on all the nodes that are at most $K$ steps away), where $K$ is the number of successive operations of convolutional layers in the neural network model.

In each layer, *Decagon* propagates latent node feature information across edges of the graph, while taking into account the type (relation) of an edge (Schlichtkrull *et al.*, 2017). A single layer of this neural network model takes the following form:

$$\mathbf{h}_i^{(k+1)} = \phi\left(\sum_r \sum_{j \in \mathcal{N}_r^i} c_r^{ij} \mathbf{W}_r^{(k)} \mathbf{h}_j^{(k)} + c_r^i \mathbf{h}_i^{(k)}\right), \quad (1)$$

where $\mathbf{h}_i^{(k)} \in \mathbb{R}^{d(k)}$ is the hidden state of node $v_i$ in the $k$-th layer of the neural network with $d^{(k)}$ being the dimensionality of this layer's representation, $r$ is a relation type, and matrix $\mathbf{W}_r^{(k)}$ is a relation-type specific parameter matrix. Here, $\phi$ denotes an non-linear element-wise activation function (*i.e.*, a rectified linear unit), which transforms the representations to be used in the layer of the neural model, $c_r^{ij}$ and $c_r^i$ are normalization constants, which we choose to be symmetric $c_r^{ij} = 1/\sqrt{|\mathcal{N}_r^i||\mathcal{N}_r^j|}$ and $c_r^i = 1/|\mathcal{N}_r^i|$ with $\mathcal{N}_r^i$ denoting the set of neighbors of node $v_i$ under relation $r$. Importantly note that the sum in Eq. 1 ranges only over the neighbors $\mathcal{N}_r^i$ of a given node $i$ and thus the computational architecture (*i.e.*, the neural network) is different for every node. Figure 3A shows an example of a per-layer convolutional update Eq. (1) for node $C$ from Figure 1. And, Figure 3C then illustrates that different nodes have different structures of neural networks (because each node's network neighborhood is different).

A deeper model can be built by chaining multiple (*i.e.*, $K$) of these layers (Figure 3A) with appropriate activation functions. To arrive at the final embedding $\mathbf{z}_i \in \mathbb{R}^d$ of node $v_i$, we compute its representation as: $\mathbf{z}_i = \mathbf{h}_i^{(K)}$. The overall encoder then takes the following form. We stack $K$ layers as defined in Eq. (1) such that the output of the previous layer becomes the input to the next layer. The input to the first layer are node feature vectors, $\mathbf{h}_i^{(0)} = \mathbf{x}_i$, or unique one-hot vectors for every node in the graph if no features are present.

### 4.2 Tensor factorization decoder

So far, we introduced *Decagon*'s encoder. The encoder maps each node $v_i \in \mathcal{V}$ to an embedding, a real-valued vector representation $\mathbf{z}_i \in \mathbb{R}^d$, where $d$ is the dimensionality of node representations. We proceed by describing the decoder component of *Decagon*.

The goal of decoder is to reconstruct labeled edges in $G$ by relying on learned node embeddings and by treating each label (edge type) differently. In particular, decoder scores a $(v_i, r, v_j)$-triple through a function $g$ whose goal is to assign a score $g(v_i, r, v_j)$ representing how likely it is that drugs $v_i$ and $v_j$ are interacting through a relation/side effect type $r$ (Figure 3B). Using embeddings for nodes $i$ and $j$ returned by *Decagon*'s encoder (Section 4.1) $\mathbf{z}_i$ and $\mathbf{z}_j$, the decoder predicts a candidate edge $(v_i, r, v_j)$ through a factorized operation:

$$g(v_i, r, v_j) = \begin{cases} \mathbf{z}_i^T \mathbf{D}_r \mathbf{R} \mathbf{D}_r \mathbf{z}_j & \text{if } v_i \text{ and } v_j \text{ are drugs} \\ \mathbf{z}_i^T \mathbf{M}_r \mathbf{z}_j & \text{if } v_i \text{ and } v_j \text{ are both proteins, or,} \\ & v_i \text{ and } v_j \text{ are a protein and a drug} \end{cases} \quad (2)$$

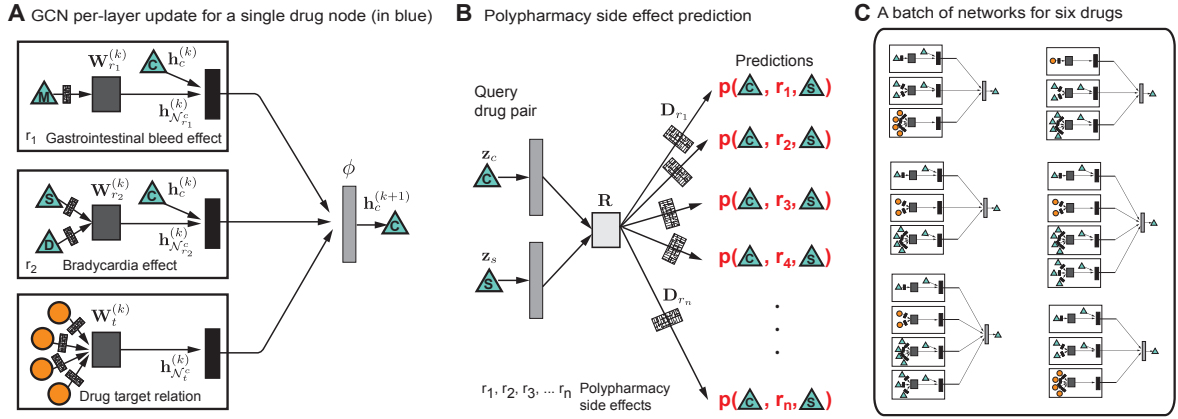

**Fig. 3.** Overview of Decagon model architecture. (A) An Decagon encoder. Shown is a per-layer update for a single graph node (a drug node representing Ciprofloxacin based on the small example input graph in Figure 1). Hidden state activations from neighboring nodes $\mathcal{N}_r^{v_c}$ are gathered and then transformed for each relation type $r$ individually (i.e., gastrointestinal bleed, bradycardia, and drug target relation). The resulting representation is accumulated in a (normalized) sum and passed through a non-linear activation function (i.e., ReLU) to produce hidden state of node $v_c$ in the $(k + 1)$-th layer, $\mathbf{h}_i^{(k+1)}$. This per-node update is computed in parallel with shared parameters across the whole graph. (B) For every relation, Decagon decoder takes pairs of embeddings (e.g., hidden node representations $\mathbf{z}_c$ and $\mathbf{z}_s$ representing Ciprofloxacin and Simvastatin) and produces a score for every (potential) edge in the graph. Shown is the decoder for polypharmacy side effects relation types. (C) A batch of neural networks that compute embeddings of six drug nodes in the input graph. In Decagon, neural networks differ from node to node but they all share the same set of relation-specific trainable parameters (i.e., the parameters of the encoder and decoder; see Eq. (1) and Eq. (2)). That is, rectangles with the same shading patterns share parameters, and thin rectangles with black and white shading pattern denote densely-connected neural layers.

followed by the application of a sigmoid function $\sigma$ to compute probability of edge $(v_i, r, v_j)$:

$$p_r^{ij} = p((v_i, r, v_j) \in \mathcal{R}) = \sigma(g(v_i, r, v_i)).\qquad(3)$$

Next, we explain *Decagon*'s decoder by distinguishing between the following two cases:

(1) When $v_i$ and $v_j$ are drug nodes, the decoder $g$ in Eq. (2) assumes a global model of drug-drug interactions (i.e., $\mathbf{R}$) whose variation and importance across polypharmacy side effects are described by side-effect-specific diagonal factors (i.e., $\mathbf{D}_r$). Here, $\mathbf{R}$ is a trainable parameter matrix of shape $d \times d$ that models global drug-drug interactions across all possible polypharmacy side effects. Additionally, in *Decagon*, every relation $r$ representing a different polypharmacy side effect is associated with a diagonal $d \times d$ matrix $\mathbf{D}_r$ modeling the importance of each dimension in $\mathbf{z}_i$ towards side effect $r$. In an alternative view, this decoder can be thought of as a tensor factorization (more specifically, a rank-$d$ DEDICOM tensor decomposition (Nickel *et al.*, 2011; Trouillon *et al.*, 2016)) of a three-way tensor, where two modes are identically formed by the drugs and the third mode holds polypharmacy side effects of drug combinations. However, a distinguishing characteristic of *Decagon* is the reliance on the encoder. Whereas classic tensor decompositions use node representations optimized directly in training, we compute them in an end-to-end fashion where node embeddings are optimized jointly together with the tensor factorization.

(2) When $v_i$ and $v_j$ are not both drug nodes, the decoder $g$ in Eq. (2) employs a bilinear form to decode edges from node embeddings. More precisely, in that case, the decoding function $g$ is associated with a trainable parameter matrix $\mathbf{M}_r$ of shape $d \times d$ that models interactions between every two dimensions in $\mathbf{z}_i$ and $\mathbf{z}_j$. The predicted edge probability is then computed using a bilinear form (Eq. (2)) followed by the application of a sigmoid function (Eq. (3)).

The use of different edge decoders based on the type of nodes in Eq. (2) is crucial because of the following two reasons: First, *Decagon* decoder can be seen as a form of effective parameter sharing between different relation types. In particular, relation types involving drug pairs use the same global drug-drug interaction model (i.e., matrix $\mathbf{R}$) containing patterns that hold true across all drug-related relation types. We expect that this decoding parameterization can alleviate overfitting on rare side effects as parameters are shared between both rare (*e.g.*, myringitis or nasal polyps) and frequent (*e.g.*, hypotension or anaemia) side effects. Second, we want a high score $g(v_i, r, v_j)$ to indicate an association between a drug combination $(v_i, v_j)$ and a side effect $r$ that cannot be attributed to $v_i$ or $v_j$ alone. To capture the polypharmacy combinatorics (Jia *et al.*, 2009), it is thus important that *Decagon* allows, through $\mathbf{R}$, for a non-zero interaction between any two dimensions in $i$'s and $j$'s embeddings.

Taken together, the trainable parameters of *Decagon* model are: (1) relation-type-specific neural network weight matrices $\mathbf{W}_r$, (2) relation-type-specific parameter matrices $\mathbf{M}_r$, (3) a global side-effect parameter matrix $\mathbf{R}$, and (4) side-effect-specific diagonal parameter matrices $\mathbf{D}_r$. *Decagon* encoder and decoder thus forms an end-to-end trainable model for multirelational link prediction in a multimodal graph (Figure 3).

Next we shall describe how to train the *Decagon* approach. In particular, we explain how to train neural network weights and interaction parameter matrices using an end-to-end learning technique.

### 4.3 *Decagon* model training

During model training, we optimize model parameters using the cross-entropy loss:

$$J_r(i, j) = -\log p_r^{ij} - \mathbb{E}_{n \sim P_r(j)} \log(1 - p_r^{in}),\qquad(4)$$

to encourage the model to assign higher probabilities to observed edges $(v_i, r, v_j)$ than to random non-edges. As in previous work (Mikolov *et al.*, 2013; Trouillon *et al.*, 2016), we estimate the model through negative sampling. For each drug-drug edge $(v_i, r, v_j)$ (*i.e.*, a positive example) in the graph, we sample a random edge $(v_i, r, v_n)$ (*i.e.*, a negative example) by randomly choosing node $v_n$. This is achieved by replacing node $v_j$ in edge $(v_i, r, v_j)$ with node $v_n$ that is selected randomly according to a sampling distribution $P_r$ (Mikolov *et al.*, 2013). Considering all edges, the final loss function in *Decagon* is:

$$J = \sum_{(v_i, r, v_j) \in \mathcal{R}} J_r(i, j).\qquad(5)$$

Recent results have shown that modeling graph-structured data can often be significantly improved with end-to-end learning (Defferrard *et al.*, 2016;

Gilmer *et al.*, 2017), thus we take an end-to-end optimization approach and jointly optimize over all trainable parameters and propagate loss function gradients through both *Decagon*'s encoder as well as decoder.

To optimize the model we train it for a maximum of 100 epochs (training iterations) using the Adam optimizer (Kingma and Ba, 2014) with a learning rate of 0.001 and early stopping with a window size of 2, *i.e.*, we stop training if the validation loss does not decrease for two consecutive epochs. We initialize weights using the initialization described in Glorot and Bengio (2010) and accordingly normalize node feature vectors. In order for the model to generalize well to unobserved edges we apply a regular dropout (Srivastava *et al.*, 2014) to hidden layer units (Eq. (1)). In practice, we use efficient sparse matrix multiplications, with complexity linear in the number of edges in $G$, to implement *Decagon* model.

We use mini-batching by sampling contributions to the loss function in Eq. (5). That is, we process multiple training mini-batches, each obtained by sampling only a fixed number of contributions from the sum over edges in Eq. (5), resulting in dynamic batches of computation graphs (Figure 3C). By only considering a fixed number of contributions to the loss function, we can remove respective data points that do not appear in the current mini-batch. This serves as an effective means of regularization, and reduces the memory requirement to train the model, which is necessary so that we can fit the full model into GPU memory[1].

# 5 Experimental setup

We view the problem of predicting polypharmacy side effects as solving a multirelational link prediction task. Here, every drug pair is connected through zero, one, or more relation types (*i.e.*, side effect types) from a set of all relation types (*i.e.*, all side effect types, see Section 2 and Figure 1).

For each polypharmacy side effect type, we split drug pairs associated with that side effect into training, validation, and test sets, ensuring that the validation and test sets each include 10% of drug pairs. For each side effect type, we use 80% of drug pairs to train a model, and 10% of drug pairs to select model parameters. The task is then to predict pairs of drugs that are associated with each side effect type. Note that we are extremely careful that there is information leakage between the folds and that the cross-validation is fair.

We apply *Decagon*, which for every drug pair and for every side effect type calculates a probability that a given drug pair is associated with a given side effect. Additionally, we integrate side information, *i.e.*, side effects of individual drugs (Section 2), into the model in the form of additional features $\mathbf{x}_i$ for drug nodes $i$. To prevent any circularity and information leakage in the evaluation, we make sure that: (1) side effects we are predicting over are true polypharmacy side effects (*i.e.*, a given polypharmacy side effect is only associated with a drug pair and not with any individual drug in the pair), and (2) no side effect types that we are predicting over are included in the side features. For example, nausea is one polypharmacy side effect, and we therefore remove all instances of nausea as a side effect for individual drugs. We note that this is a conservative approach which allows us to reliably estimate prediction performance.

We are not aware of any other approach developed for predicting side effects of drug pairs. We thus evaluate the performance of *Decagon* against the following multirelational link prediction approaches:

- **RESCAL tensor decomposition** (Nickel *et al.*, 2011): This is a tensor factorization approach that takes a multirelational structure into account. Given $\mathbf{X}_i$, a drug-drug matrix encoding associations of drugs pairs with side effect $r$, matrix $\mathbf{X}_i$ is decomposed as: $\mathbf{X}_r = \mathbf{A}\mathbf{T}_r\mathbf{A}^T$ for $r = 1, 2, \ldots, 964$, where $\mathbf{T}_r$ and $\mathbf{A}$ are model parameters. Given drugs $i$ and $j$, their association with $r$ is predicted as: $\mathbf{a}_i \mathbf{T}_r \mathbf{a}_j$.

- **DEDICOM tensor decomposition** (Papalexakis *et al.*, 2017): This is a related tensor factorization approach suitable for sparse data

settings. A given drug-drug matrix $\mathbf{X}_i$ is decomposed as: $\mathbf{X}_r = \mathbf{A}\mathbf{U}_r\mathbf{T}\mathbf{U}_r\mathbf{A}^T$. Given drugs $i$ and $j$, their association with $r$ is predicted as: $\mathbf{a}_i\mathbf{U}_r\mathbf{T}\mathbf{U}_r\mathbf{a}_j$.

- **DeepWalk neural embeddings** (Perozzi *et al.*, 2014; Zong *et al.*, 2017): This approach learns $d$-dimensional neural features for nodes based on a biased random walk procedure exploring network neighborhoods of nodes. Drug pairs are represented by concatenating learned drug feature representations and used as input to a logistic regression classifier. For each link-type (*i.e.*, side effect type) we train a separate logistic regression classifier.

- **Concatenated drug features**: This approach constructs a feature vector for each drug based on PCA representation of drug-target protein interaction matrix and based on PCA representation of side effects of individual drugs. Drug pairs are represented by concatenating the corresponding drug feature vectors and used as input to a gradient boosting trees classifier that then predicts the exact side effect of a pair of drugs.

The parameter settings for every approach are determined using a validation set with a grid search over candidate parameter values (*e.g.*, for gradient boosting trees, the number of trees used was varied from 10 to 100). In case an approach is not a multirelational link prediction method, we select parameters with best performance on the validation set individually for each side effect type. Specifically, *Decagon* uses a 2-layer neural architecture with $d(1) = 64$, and $d(2) = 32$ hidden units in each layer, a dropout rate of 0.1, and a mini-batch size of 512 in all experiments.

Performance is calculated individually per side effect type using area under the receiver-operating characteristic (AUROC), area under the precision-recall curve (AUPRC), and average precision at 50 (AP@50). Higher values always indicate better performance.

# 6 Results

*Decagon* operates on multimodal graphs and in highly multirelational settings. This flexibility makes *Decagon* especially suitable for predicting side effects of pairs of drugs as we shall discuss below.

## 6.1 Prediction of polypharmacy side effects

We start by comparing the performance of *Decagon* to alternative approaches. From results in Table 2, we see that considering the multimodal network representation and modeling a large number of different side effects allows *Decagon* to outperform other approaches by a large margin. Across 964 side effect types, *Decagon* outperforms alternative approaches by 19.7% (AUROC), 22.0% (AUPRC), and 36.3% (AP@50). *Decagon*'s improvement is especially pronounced relative to tensor factorization methods, where *Decagon* surpasses tensor-based methods by up to 68.7% (AP@50). This finding highlights a potential limitation of directly optimizing a tensor decomposition (*i.e.*, vanilla RESCAL and DEDICOM (Nickel *et al.*, 2011; Papalexakis *et al.*, 2017)) without relying on a graph-structured convolutional encoder. We also compared *Decagon* with two other methods (Perozzi *et al.*, 2014; Zong *et al.*, 2017), which we adapted for a multirelational link prediction task. We observe that DeepWalk neural embeddings and Concatenated drug features achieve a gain of 9.0% (AUROC) and a 20.1% gain (AUPRC) over tensor-based methods. However, these approaches employ a two-stage pipeline, consisting of a drug feature extraction model and a link prediction model, both of which are trained separately. Furthermore, they cannot consider interdependence of different side effects that we showed to contain useful information (Section 3). These additional modeling insights, give *Decagon* a 22.0% gain over DeepWalk neural embeddings, and a 12.8% gain over Concatenated drug features in AP@50 scores.

---

[1] All data and code are released on the project website.

Table 2. Area under ROC curve (AUROC), area under precision-recall curve (AUPRC), and average precision at 50 (AP@50) for polypharmacy side effect prediction. Reported are average performance values for 964 side effect types.

| Approach | AUROC | AUPRC | AP@50 |
|---|---|---|---|
| *Decagon* | 0.872 | 0.832 | 0.803 |
| RESCAL tensor factorization | 0.693 | 0.613 | 0.476 |
| DEDICOM tensor factorization | 0.705 | 0.637 | 0.567 |
| DeepWalk neural embeddings | 0.761 | 0.737 | 0.658 |
| Concatenated drug features | 0.793 | 0.764 | 0.712 |

Table 3. Side effects with the best and worst performance in Decagon.

| Best performing side effects | AUPRC | Worst performing side effects | AUPRC |
|---|---|---|---|
| Mumps | 0.964 | Bleeding | 0.679 |
| Carbuncle | 0.949 | Increased body temperature | 0.680 |
| Coccydynia | 0.943 | Emesis | 0.693 |
| Tympanic membrane perfor. | 0.941 | Renal disorder | 0.694 |
| Dyshidrosis | 0.938 | Leucopenia | 0.695 |
| Spondylosis | 0.929 | Diarrhea | 0.705 |
| Schizoaffective disorder | 0.919 | Icterus | 0.707 |
| Breast dysplasia | 0.918 | Nausea | 0.711 |
| Ganglion | 0.909 | Itch | 0.712 |
| Uterine polyp | 0.908 | Anaemia | 0.712 |

These findings are aligned with results that predictions can often be significantly improved by end-to-end learning and specifically using graph auto-encoders (Kipf and Welling, 2016; Hamilton *et al.*, 2017a,b). In particular, tensor decomposition and neural embedding baseline approaches allow us to quantify what percentage of the performance improvement is due to the embeddings (*i.e.*, *Decagon*'s encoder) and what percentage is due to the multitask learning (*i.e.*, *Decagon*'s decoder).

To better understand *Decagon*'s performance we stratify the aggregated statistics in Table 2 by side effect type. Manual examination of the results and a discussion with domain experts reveals a common property of best performing side effects in Table 3. We observe that *Decagon* models particularly well side effects with strong apparent molecular underpinnings. This observation is consistent with our expectation because *Decagon*'s multimodal graph (Figure 1) contains predominantly pharmacogenomic information. We also observed that side effects with the worst performance tend to be common side effects and/or have non-molecular origins with potentially important environmental and behavioral components (Table 3). *Decagon*'s competitive performance on those side effects can be explained by effective sharing of model parameters across different types of side effects.

### 6.2 Investigation of *Decagon*'s novel predictions

Next, we perform a literature-based evaluation of new hits. Our goal is to evaluate the quality of novel *Decagon*'s predictions about relationships between side effects and drug pairs. To this aim, we ask *Decagon* to make a prediction for every drug pair and every side effect type in the dataset. We then use these predictions to construct a ranked list of (drug $i$, side effect type $r$, drug $j$) triples, where the triples are ranked by predicted probability scores $p_r^{ij}$ (Eq. (3)). We then exclude from the ranked list all the known associations between drug pairs and side effects, and afterwards investigate the ten highest ranked predictions in the list. To prevent the risk of investigative bias we do not allow any crosstalk between different stages of the analysis. We then search published literature to see if we can find supporting evidence for these novel predictions.

Table 4 shows *Decagon*'s predictions and literature evidence supporting these predictions. We were able to find literature evidence for five out of ten highest ranked predicted side effects. That is, our method both correctly identified the drug pair as well as the side effect type for these highest ranked predictions. This result is remarkable because the predictions were specific and the supporting evidence was very unlikely to be found by random selection of drug pair and side effect associations.

Table 4. New polypharmacy side effect predictions given by (drug $i$, side effect type $r$, drug $j$) triples that were assigned the highest probability scores by Decagon. For each prediction, we include its rank $k$ in the ranked list of all predictions and literature evidence supporting existence of the predicted association.

| $k$ | Polypharmacy effect $r$ | Drug $i$ | Drug $j$ | Evidence |
|---|---|---|---|---|
| 1 | Sarcoma | Pyrimethamine | Aliskiren | Stage *et al.* |
| 4 | Breast disorder | Tolcapone | Pyrimethamine | Bicker *et al.* |
| 6 | Renal tubular acidosis | Omeprazole | Amoxicillin | Russo *et al.* |
| 8 | Muscle inflammation | Atorvastatin | Amlodipine | Banakh *et al.* |
| 9 | Breast inflammation | Aliskiren | Tioconazole | Parving *et al.* |

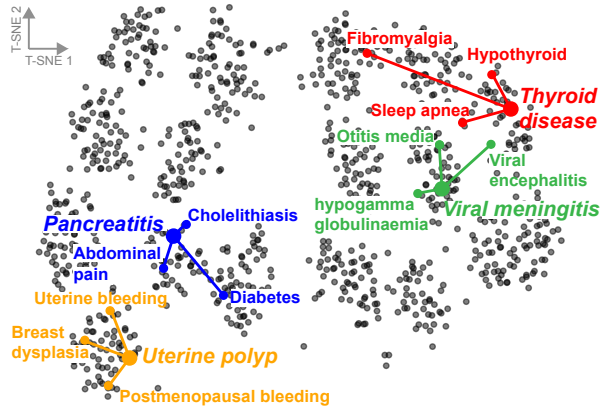

**Fig. 4.** Visualization of side effects in Decagon. The side effects are mapped to the 2D space using the t-SNE package (Maaten and Hinton, 2008) with learned side effect representations ($D_r$, $r = 1, 2, \ldots, 964$, see Eq. (2)) as input. Selected side effects are uterine polyp, pancreatitis, viral meningitis, and thyroid disease. For each selected side effect, we highlight three side effects that most often co-occur with the selected side effect in the drug combination dataset.

We note that the cited literature explicitly investigates interactions between the predicted drug pair and the predicted side effect. For example, *Decagon* signified the use Atorvastatin and Amlodipine can lead to muscle inflammation (Table 4, 8th highest ranked prediction). In fact, recent reports (*e.g.*, Banakh *et al.* (2017)) have found injuries in muscle tissue due to presumed drug interactions of Atorvastatin with Amlodipine. *Decagon* also flagged a potential association between Pyrimethamine, an antimicrobial that, if taken alone, is effective in the treatment of malaria, and Aliskiren, a renin inhibitor, whose clinical trial was halted after discovered kidney complications (Parving *et al.*, 2012), suggesting an increased risk of cancer (1st highest ranked prediction). The analysis here demonstrates the potential of *Decagon*'s predictions to facilitate the translational science and the discovery of novel (non-)efficacious drug combinations.

### 6.3 Exploration of *Decagon*'s side effect embeddings

Finally, we are interested in knowing whether *Decagon* meets the design goals presented in Section 3. In particular, we test if *Decagon* can capture the interdependence of different side effect types revealed by our exploratory data analysis (2nd observation in Section 3). To this aim, we take diagonal matrices $D_r$, which specifically model the importance of interactions for each side effect type $r$ in *Decagon*'s multirelational link prediction (Section 4.2). We extract the diagonal from each $D_r$ and use it as a vector representation for side effect $r$. We embed these vector representations into a 2D space using t-SNE (Maaten and Hinton, 2008) and then visualize in Figure 4.

Figure 4 reveals the existence of clustering structure in side effects' representations. Examining the figure, we observe that side effects embedded close together in the 2D space tend to co-occur in drug combinations. This observation indicates that *Decagon* infers similar matrices $D_{r_1}$ and $D_{r_2}$ for side effects $r_1$ and $r_2$ that appear together in many drug combinations. For example, the top three side effects that often appear together with uterine polyp side effect are: uterine bleeding, breast dysplasia, and postmenopausal bleeding. Indeed, *Decagon* infers similar diagonal factors $D_r$ for all three side effects, resulting in localized projections in the 2D space (Figure 4).

To test if the appealing pattern in Figure 4 holds true across many side effect types we proceed as follows. We compute average Euclidean distance between each side effect's vector representation and vector representations of three most frequently co-occurring side effects. We find that co-occurring/related side effects have significantly more similar representations (*i.e.*, diagonal factors $D_r$) than expected by chance (p-value = $1e{-}34$, 2-sample KS test). We thus conclude that *Decagon* is able to meet the design goals of polypharmacy side effect modeling. Furthermore, the analysis here indicates that *Decagon*'s multirelational link prediction model (Section 4.2) can capture interdependence of side effects present in drug combination data.

## 7 Related work

We review related research on computational prediction of drug combinations, and on neural networks for graph-structured data.

**Drug combination modeling.** Methods in computational pharmacology aim to find associations between drugs and molecular targets, predict potential adverse drug reactions, and find new uses of existing drugs (Campillos *et al.*, 2008; Li *et al.*, 2015; Hodos *et al.*, 2016). In contrast to individual drugs and single drug therapy (*i.e.*, monotherapy) predominantly considered by these methods, we consider drug combinations (*i.e.*, polypharmacy). This is important as polypharmacy is a useful strategy for combating complex diseases (Jia *et al.*, 2009; Han *et al.*, 2017) with important implications for health care system (Ernst and Grizzle, 2001).

Traditionally, effective drug combinations have been identified by experimentally screening all possible combinations of a pre-defined set of drugs (Chen *et al.*, 2016b). Given the large number of drugs, experimental screens of pairwise combinations of drugs pose a formidable challenge in terms of cost and time. For example, given $n$ drugs, there are $n(n-1)/2$ pairwise drug combinations and many more higher-order combinations. To address the combinatorial explosion of candidate drug pairs that potentially interact, computational methods were developed to identify drug pairs that potentially interact, *i.e.*, drug pairs that produce an exaggerated response over and beyond the additive response expected under no interaction (Ryall and Tan, 2015). Previous research in this realm focused on defining drug-drug interactions through the concepts of synergy and antagonism (Loewe, 1953; Lewis *et al.*, 2015), quantitatively measuring dose-effect curves (Bansal *et al.*, 2014; Takeda *et al.*, 2017), and determining whether or not a given drug pair interacts according to an experiment measuring cell viability (Huang *et al.*, 2014b,a; Sun *et al.*, 2015; Zitnik and Zupan, 2016; Chen *et al.*, 2016b,a; Shi *et al.*, 2017). All of these approaches predict drug-drug interactions as scalar values representing the overall probability/strength of an interaction for a given drug pair. In sharp contrast, our work here goes a step further and identifies how exactly, if at all, a given drug pair manifests clinically within a patient population. In particular, we model clinical manifestations that cannot be attributed to either drug alone and that arise due to drug interaction (*i.e.*, polypharmacy side effects). Whereas previous research focused on generating pointwise interaction estimates representing cell viability or a closely related outcome in an experimental drug screen, we predict, for the first time, which, if any, polypharmacy side effects can occur when multiples drugs are taken together by a patient, yielding a more direct path for clinical translation.

Although present drug-drug interaction prediction approaches cannot be directly used for the problem studied here, we briefly overview methodology used by these approaches. Drug-drug interaction prediction approaches can be categorized into classification-based and similarity-based methods. Classification-based methods consider drug-drug interaction prediction as a binary classification problem (Cheng and Zhao, 2014; Huang *et al.*, 2014a; Zitnik and Zupan, 2016; Chen *et al.*, 2016b; Shi *et al.*, 2017). These methods use known interacting drug pairs as positive examples and other drug pairs as negative examples, and train classification models, such as naïve Bayes, logistic regression, and support vector machine. In contrast, similarity-based methods assume that similar drugs may have similar interaction patterns (Gottlieb *et al.*, 2012; Vilar *et al.*, 2012; Huang *et al.*, 2014b; Li *et al.*, 2015; Zitnik and Zupan, 2015; Sun *et al.*, 2015; Li *et al.*, 2017). These methods use different kinds of drug-drug similarity measures defined on drug chemical substructures, interaction profile fingerprints, drug side effects, off-side effects, and connectivity of molecular targets. The methods aggregate similarity measures through clustering or label propagation in order to identify potential drug-drug interactions (Zhang *et al.*, 2015; Ferdousi *et al.*, 2017; Zhang *et al.*, 2017). However, all these methods generate drug-drug interaction scores and do not predict the exact polypharmacy side effect, which is the goal of our work here.

**Neural networks on graphs.** Our model extends existing work in the field of neural networks on graphs (Hamilton *et al.*, 2017b; Kipf and Welling, 2016; Defferrard *et al.*, 2016; Hamilton *et al.*, 2017a; Schlichtkrull *et al.*, 2017; Gilmer *et al.*, 2017). Neural networks on graphs enable learning over graph structures by generalizing the notion of convolution operation typically applied to image datasets to operations that can operate on arbitrary graphs. These neural networks can also be seen as an embedding methodology that distills high-dimensional information about each node's neighborhood into a dense vector embedding without requiring manual feature engineering. In particular, graph convolutional networks (Kipf and Welling, 2016; Defferrard *et al.*, 2016; Hamilton *et al.*, 2017a) and message passing neural networks (Gilmer *et al.*, 2017) are related lines of research that allow for layer-wise learning of node embeddings in graphs.

Although graph convolutional networks achieve state-of-the-art performance on important prediction problems in social networks and knowledge graphs, they have not yet been used for problems in computational biology. Our model extends graph convolutional networks by incorporating support for multiple edge types, each type representing a different side effect, and by providing a form of efficient weight sharing for multimodal graphs with a large number of edge types.

## 8 Conclusion

We presented *Decagon*, an approach for predicting side effects of drug pairs. *Decagon* is a general graph convolutional neural network designed to operate on a large multimodal graph where nodes can be connected through a large number of different relation types. We use *Decagon* to, for the first time, infer a prediction model that can identify side effects of pairs of drugs. *Decagon* predicts an association between a side effect and a co-prescribed drug pair (*i.e.*, a drug combination) to identify side effects that cannot be attributed to either drug alone. The graph convolutional model achieves excellent accuracy on the polypharmacy side effect prediction task, allows us to consider nearly a thousand different side effect types integrating molecular and patient population data, and provides insights into clinical manifestation of drug-drug interactions.

There are several directions for future work. Our approach integrates molecular protein-protein and drug-target networks together with population-level patients' side effect data. Other sources of biomedical information, such as dosed concentration levels of drugs, might be relevant for modeling side effects of drug pairs, and we hope to investigate the utility of integrating them into the model. As *Decagon*'s graph convolutional model is a general approach for multirelational link prediction in any multimodal network, it would be interesting to apply it to other domains and problems, for example, finding associations between patient outcomes and comorbid diseases, or for identifying dependencies between mutant phenotypes and gene-gene interactions.

# Funding

This research has been supported in part by NSF, NIH BD2K, DARPA SIMPLEX, Stanford Data Science Initiative, and Chan Zuckerberg Biohub. *Conflict of Interest:* none declared.